\documentclass[lettersize,journal]{IEEEtran}
\usepackage{amsmath,amsfonts}
\usepackage{algorithmic}
\usepackage{algorithm}
\usepackage{array}
\usepackage{bm}
\usepackage[caption=false,font=normalsize,labelfont=sf,textfont=sf]{subfig}
\usepackage{textcomp}
\usepackage{stfloats}
\usepackage{url}
\usepackage{verbatim}
\usepackage{graphicx}
\usepackage{booktabs}
\usepackage{siunitx}
\usepackage{float}
\usepackage[hidelinks]{hyperref}
\usepackage[
    backend=biber,
    style=ieee,
    sorting=none,
    doi=false,
    url=false,
    isbn=false,
    maxnames=6,
    minnames=1
]{biblatex}
\usepackage{balance}
\newcommand{\Rtwo}{$R^2$}
\newcommand{\modelname}{SATURN}

\usepackage{fancyhdr}
\emergencystretch=\maxdimen
\begin{document}

\title{\textbf{Temporal Graph Learning of Wearable Actigraphy and Sleep Traces for Modelling Adolescent Crystallized Intelligence}}

\author{
Md. Tanvir Rahman,
Nabil Anan Orka,
Asaduzzaman Khan, and
Mohammad Ali Moni
\thanks{Md. Tanvir Rahman, Nabil Anan Orka, Asaduzzaman Khan, and Mohammad Ali Moni are with the School of Health and Rehabilitation Sciences, The University of Queensland, QLD 4072, Australia.}
\thanks{Md. Tanvir Rahman is also with the Department of Information and Communication Technology, Mawlana Bhashani Science and Technology University, Tangail 1902, Bangladesh.}
}

\markboth{IEEE Transactions on Computational Social Systems}
{Rahman \MakeLowercase{\textit{et al.}}: SATURN for Wearable-Based Cognitive Phenotyping}

\maketitle
\thispagestyle{fancy} 

\begin{abstract}
Wearable actigraphy offers a scalable, ecologically valid alternative to episodic clinical assessment. However, predicting continuous adolescent crystallized intelligence ($G_c$) from such traces remains challenging due to irregular device adherence and complex behavioral-environmental interactions. We address this using daily summary data derived from 21-day Fitbit records of 6,091 adolescents in the Adolescent Brain Cognitive Development Study (Release 5.1). We propose SATURN, a Sleep-Activity Temporal Unified Regression Network. It represents participants as 21-node temporal graphs encoding daily behaviors and temporal adjacency. To prevent imputation artifacts, invalid-day edges are dynamically pruned during forward passes. Node embeddings are refined via residual GATv2 layers, aggregated through masked attention pooling, and fused with sociodemographic covariates. Under family-controlled, age-sex-BMI-stratified cross-validation, SATURN achieves $R^2 = 0.2783 \pm 0.0127$, consistently improving upon flattened machine learning (Gradient Boosting, $R^2 = 0.2372$) and sequential deep learning (BiLSTM, $R^2 = 0.2688$) baselines. Explainability analyses identify light activity, metabolic equivalents, and sleep duration as dominant predictors, while Monte Carlo dropout and subgroup analyses confirm equitable performance across sociodemographic strata. Ultimately, SATURN establishes a rigorous computational framework for digital cognitive phenotyping, offering a scalable pathway to complement traditional assessments by highlighting macro-level behavioral anomalies.
\end{abstract}


\section{Introduction}
\IEEEPARstart{W}{earable} computing systems enable continuous, unobtrusive monitoring of everyday behavior at a scale and ecological fidelity that traditional clinic-based assessments fundamentally cannot replicate \cite{srikrishnarka2024wearable}. Commercial devices such as Fitbit now generate longitudinal records of physical activity, sleep architecture, heart rate, and sedentary behavior across natural environments \cite{master2022association,zheng2024sleep}, enabling a shift from episodic to persistent cognitive monitoring. Recent advances in computational social systems have demonstrated the value of large-scale human--device data for population-level mental health surveillance and predictive healthcare \cite{zhou2021detecting, Xu2022, baucas2023federated}, while deep learning has transformed neurocognitive modeling from neuroimaging data \cite{rahman2025understanding}. The convergence of these trends raises a tractable but underexplored question: \textit{do multi-week consumer wearable traces contain a reliable, continuous signal for standardized cognitive ability in adolescents?}

In this study, we focus on crystallized intelligence ($G_c$), the NIH Toolbox construct that reflects accumulated language, reading, and knowledge-based skills \cite{gershon2013nih, weintraub2013cognition, luciana2018abcd}, for two reinforcing reasons. First, $G_c$ is shaped by the same behavioral ecology that wearables capture. It develops gradually through daily routines, family contexts, and structured opportunity \cite{hackman2009ses, loughnan2023intelligence, marzoratti2026contextualizing}, making it a natural target for longitudinal behavioral fingerprinting. Second, contemporary movement-behavior research establishes that physical activity, sedentary time, and sleep operate as an integrated system rather than independent channels, and that their combined variation predicts cognitive and academic outcomes \cite{wilhite2023combinations, yang2022effects}. Together, these properties suggest that a sufficient wearable trace should encode a $G_c$-relevant signal, but extracting it requires solving three structural problems simultaneously.

The first is temporal irregularity. Free-living wearable streams are incompletely observed. Device non-wear, school schedules, and family routines produce gaps that are not missing at random but carry behavioral meaning \cite{che2018recurrent, kiang2021sociodemographic, currey2023increasing}. Standard architectures that treat absent days as zero-activity observations conflate measurement failure with genuine sedentary behavior, corrupting the signal before modeling begins. The second is behavioral integration. Sleep and activity are jointly informative. Their relationship across days encodes more than either channel in isolation. Sequence models can capture temporal dependencies but do not natively represent the cross-channel, day-to-day relational structure that characterizes behavioral ecology. The third is demographic confounding. Wearable adherence, activity patterns, and $G_c$ all co-vary with sociodemographic factors and data-collection site \cite{goldsack2020verification, kim2023jama, nagata2023pediatric, yfantidou2026unfair}. A model that fails to separate these structural sources from the wearable-derived behavioral signal will inherit demographic bias rather than capture genuine cognition.

These three problems jointly motivate a graph-based representation. By encoding each observation window as a temporal graph--days as nodes, behavioral similarity as edges--we can natively represent cross-channel relational structure, apply dynamic pruning to invalid nodes before message passing, and enforce the distinction between informative absence and true low activity. We build this representation into the Sleep-Activity Temporal Unified Regression Network (\modelname{}), which augments the temporal graph with learnable distance decay, masked attention pooling, and explicit sociodemographic fusion to isolate the incremental wearable-derived signal from demographic and site-related variation. We evaluate \modelname{} on the Adolescent Brain Cognitive Development (ABCD) Study \cite{volkow2018abcd, garavan2018recruiting} using family-controlled stratified cross-validation, with subgroup-robustness and algorithmic-fairness analyses across sociodemographic strata.

Ultimately, the primary aim of this study is to establish missingness-aware temporal networks as a reliable framework for extracting continuous cognitive-behavioral signals from large-scale consumer wearables. By resolving the structural challenges of temporal irregularity and demographic confounding, this work contributes \modelname{}—a novel sociotechnical modeling approach that successfully recovers behavioral hierarchies predictive of adolescent $G_c$. Furthermore, we contribute empirical evidence demonstrating that this dynamic topological approach isolates genuine behavioral signals more effectively than conventional flattened or sequential imputation methods. Finally, by mapping model internals and epistemic uncertainty, we provide a transparent, equitable, and interpretable foundation for digital cognitive phenotyping across diverse sociodemographic strata.

\begin{figure*}[!t]
    \centering
    \includegraphics[width=\textwidth]{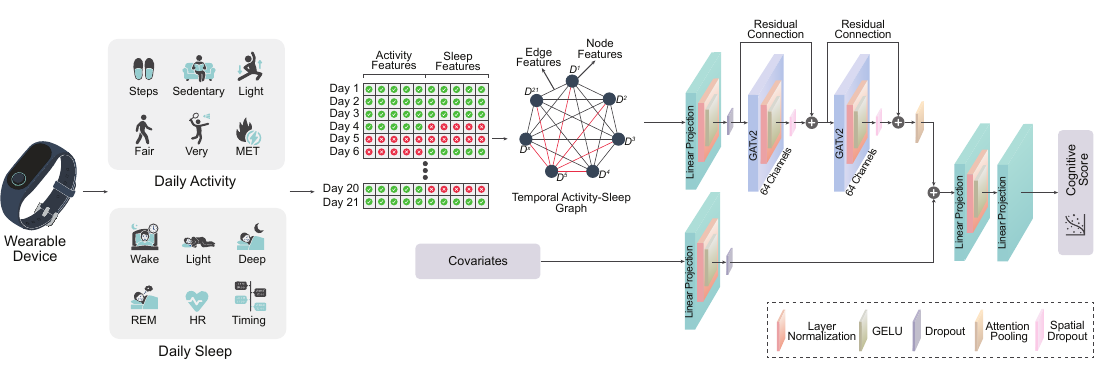}
    \caption{Structural pipeline of the proposed \modelname{} architecture. Fitbit activity and sleep records are mapped to a fixed 21-day calendar with explicit validity indicators. Each day is represented as a graph node, and the fully connected temporal base graph contains 441 directed edges with five temporal edge attributes. Invalid days are pruned from the participant-specific effective edge set during GATv2 message passing. The masked temporal representation is refined by two residual GATv2 layers, summarized by learned attention pooling, fused with encoded sociodemographic covariates, and passed to a regression head for crystallized intelligence prediction.}
    \label{fig:methodology}
\end{figure*}

\section{Materials and Methods}
We developed a multi-stage pipeline for the leakage-free prediction of adolescent crystallized intelligence from multimodal wearable activity and sleep sequences. An overview of the proposed \modelname{} architecture is provided in Fig. \ref{fig:methodology}. Reading from left to right, the pipeline extracts daily behavioral streams, maps them to a fixed 21-day calendar grid, and constructs a participant-specific temporal graph. Crucially, the schematic highlights two key implementation details: residual GATv2 message passing is applied exclusively over valid-day edges, and the sociodemographic covariate branch is fused after graph-level temporal pooling rather than being injected at every daily node.

\subsection{Data Source, Target, and Covariates}
Data were drawn from the ABCD Study Tabulated Data Release 5.1 \cite{volkow2018abcd,garavan2018recruiting}. The inputs comprised daily Fitbit-derived activity and sleep records collected at the two-year follow-up wave, mapped against the uncorrected NIH Toolbox crystallized cognition score ($G_c$). To evaluate whether wearable traces provide information beyond measured demographic and developmental structures, we included six baseline covariates: age, biological sex, BMI \cite{sakib2023adiposity}, study site, highest parental education, and combined household income \cite{hackman2009ses,loughnan2023intelligence,marzoratti2026contextualizing}. Integrating these covariates isolated the incremental behavioral signal from known anthropometric, socioeconomic, and multi-site structural variations.

\subsection{Calendarization and Daily Wearable Features}
For each participant, day 0 was defined as the start date of the Fitbit protocol. A fixed 21-day calendar was then generated from day 0 to day 20, and each calendar position was populated with the available activity record, sleep record, or both. This ensured that all participants were represented over the same intended observation window while preserving the distinction between valid observations and missing wearable data. Activity observations were retained when they corresponded to protocol-compliant wear days within the 21-day window, and sleep observations were retained when the corresponding sleep date fell within the same window. Participants were included if they had at least seven valid activity days or at least seven valid sleep days, allowing the cohort to retain adolescents with usable sleep information even when activity adherence was incomplete. The final modeling cohort contained 6,091 adolescents.

The 21-day window was treated as a protocol-aligned calendar scaffold rather than a model-tuned sequence length. This choice follows the 21-day ABCD Fitbit observation structure described in prior ABCD wearable analyses \cite{kim2023jama,nagata2023pediatric} and provides a longer monitoring window than the minimum multi-day durations often reported as sufficient for reliable youth activity and sleep estimates \cite{trost2000monitoring,antczak2021reliability}. Each calendar day was represented by contextual, activity, and sleep features. Contextual variables encoded calendar position, week number, day of week, and weekend status. Activity features summarized total and daytime-only movement, including steps, average metabolic equivalents, sedentary duration, light activity, fairly active duration, and very active duration. Sleep features summarized sleep-period duration, wake-after-sleep-onset duration, light sleep, deep sleep, rapid-eye-movement sleep, awakening count, sleep-stage heart-rate summaries, and sleep timing variables.

\subsection{Validity Masks, Encoding, and Missing-Value Handling}
Validity indicators were generated before missing values were filled. Separate binary indicators captured total-day activity validity, daytime activity validity, sleep-feature validity, and activity wear quality. Specifically, activity wear quality was flagged as valid only if daytime, nighttime, and total wear duration metrics were entirely present (non-missing), while sleep validity required the presence of core sleep-stage features. For graph construction, the final day-level validity mask was set to valid if a participant possessed complete, usable information for either the activity wear-quality stream or the sleep stream.

Calendar and clock variables were transformed into cyclic encodings before model fitting. Day of week used a seven-day sine and cosine transform, sequence day used a 21-day sine and cosine transform, week number was one-hot encoded for weeks 0--3, and sleep timing variables used a 24-hour sine and cosine transform:
\begin{equation}
    x_{\sin} = \sin\left(\frac{2\pi t}{T}\right), \qquad
    x_{\cos} = \cos\left(\frac{2\pi t}{T}\right),
\end{equation}
where \(T\) denotes the corresponding temporal period. After these transformations, the model used 47 temporal node features. The graph-validity mask was appended as an additional input channel, resulting in 48 node-level input channels per day.


To prevent data leakage, all standardization parameters were estimated only from the training subjects within each fold. Continuous temporal features were standardized using only valid training nodes. Cyclic encodings, one-hot week indicators, and validity masks were not standardized. The remaining missing feature values were set to zero, corresponding to the training-set mean for standardized continuous features and remaining distinguishable from true observations through the explicit validity mask.

\begin{table*}[t]
\caption{Demographic and Crystallized Intelligence Score Summary Across Five Family-Aware Cross-Validation Folds (Mean \(\pm\) SD)}
\label{tab:demographics}
\centering
\begin{tabular*}{\textwidth}{@{\extracolsep{\fill}} l l l l l l l l l}
\toprule
\textbf{Fold} & \textbf{Split} & \textbf{Subjects} & \textbf{Families} & \textbf{Males} & \textbf{Females} & \textbf{Age (months)} & \textbf{BMI} & \textbf{\bm{$G_c$}} \\
\midrule
1 & Train & 3904 & 3331 & 1976 & 1928 & 143.32 $\pm$ 7.78 & 20.48 $\pm$ 4.80 & 91.28 $\pm$ 6.93 \\
 & Validation & 967 & 832 & 521 & 446 & 143.27 $\pm$ 7.68 & 20.37 $\pm$ 4.74 & 91.21 $\pm$ 7.09 \\
 & Test & 1220 & 1040 & 637 & 583 & 143.51 $\pm$ 7.95 & 20.35 $\pm$ 4.79 & 91.24 $\pm$ 6.88 \\
\midrule
2 & Train & 3907 & 3330 & 2020 & 1887 & 143.36 $\pm$ 7.79 & 20.34 $\pm$ 4.74 & 91.31 $\pm$ 7.00 \\
 & Validation & 963 & 831 & 493 & 470 & 143.36 $\pm$ 7.70 & 20.72 $\pm$ 4.98 & 91.06 $\pm$ 6.80 \\
 & Test & 1221 & 1042 & 621 & 600 & 143.33 $\pm$ 7.88 & 20.51 $\pm$ 4.78 & 91.27 $\pm$ 6.90 \\
\midrule
3 & Train & 3905 & 3329 & 2025 & 1880 & 143.29 $\pm$ 7.85 & 20.36 $\pm$ 4.76 & 91.32 $\pm$ 6.87 \\
 & Validation & 970 & 834 & 482 & 488 & 143.61 $\pm$ 7.86 & 20.51 $\pm$ 4.50 & 91.38 $\pm$ 6.91 \\
 & Test & 1216 & 1040 & 627 & 589 & 143.33 $\pm$ 7.58 & 20.62 $\pm$ 5.09 & 90.99 $\pm$ 7.20 \\
\midrule
4 & Train & 3900 & 3331 & 1986 & 1914 & 143.32 $\pm$ 7.74 & 20.56 $\pm$ 4.91 & 91.21 $\pm$ 6.99 \\
 & Validation & 972 & 831 & 506 & 466 & 143.37 $\pm$ 8.03 & 20.00 $\pm$ 4.41 & 91.46 $\pm$ 6.75 \\
 & Test & 1219 & 1041 & 642 & 577 & 143.44 $\pm$ 7.80 & 20.38 $\pm$ 4.65 & 91.27 $\pm$ 6.95 \\
\midrule
5 & Train & 3916 & 3331 & 2012 & 1904 & 143.45 $\pm$ 7.76 & 20.46 $\pm$ 4.86 & 91.26 $\pm$ 6.97 \\
 & Validation & 960 & 832 & 515 & 445 & 143.20 $\pm$ 7.95 & 20.48 $\pm$ 4.71 & 90.94 $\pm$ 7.03 \\
 & Test & 1215 & 1040 & 607 & 608 & 143.14 $\pm$ 7.78 & 20.32 $\pm$ 4.60 & 91.53 $\pm$ 6.79 \\
\bottomrule
\end{tabular*}

\vspace{2mm}
\parbox{1\textwidth}{\small \textit{Note:} \(G_c\) denotes crystallized intelligence. Family IDs were used to enforce family-aware cross-validation and prevent data leakage across train, validation, and test splits.}
\end{table*}

\subsection{Family-Aware Stratified Cross-Validation}

The ABCD cohort includes siblings and twins; therefore, random subject-level splitting can leak shared familial and environmental information across training and test partitions \cite{volkow2018abcd,garavan2018recruiting}. To prevent this, we used a family-aware five-fold cross-validation strategy, assigning all members of the same family to the same split. Stratification labels were formed by crossing biological sex with age and body mass index tertiles, yielding an age--sex--BMI stratification scheme. The outer loop generated five held-out test folds, and the remaining training-validation portion of each fold was further divided using the same family-aware stratification principle. The validation split was used for learning-rate scheduling, early stopping, and checkpoint selection. The validation split was used for learning-rate scheduling, early stopping, and checkpoint selection. Table \ref{tab:demographics} shows that the resulting folds were closely balanced in sample size, family count, sex distribution, age, body mass index, and \(G_c\).

\subsection{\modelname{} Architecture}

For participant \(b\), the wearable window is represented as a directed graph \(\mathcal{G}_b=(\mathcal{V}_b,\mathcal{E}_b)\) with \(N=21\) temporal nodes. Each node corresponds to one calendar day in the observation window. The base graph is fully connected and includes self-loops, giving \(21\times21=441\) possible directed edges per participant before masking. Each base edge from source day \(i\) to destination day \(j\) is assigned five temporal attributes:
\begin{equation}
\mathbf{e}_{ij} =
\left[
\frac{j-i}{20},
\frac{|j-i|}{20},
\mathbb{I}[i=j],
\mathbb{I}[|i-j|=1],
\mathbb{I}[|i-j|=7]
\right],
\label{eq:edge_features}
\end{equation}
where \(\mathbb{I}[\cdot]\) denotes the indicator function. These attributes encode signed temporal direction, normalized absolute distance, self-connection, adjacent-day relation, and weekly-cycle relation.

Let \(\mathbf{m}_b\in\{0,1\}^{21}\) denote the graph-validity mask for participant \(b\). During every forward pass, the batch-level edge list is dynamically pruned so that an edge is retained only when both its source and destination nodes are valid:
\begin{equation}
    (i,j)\in\mathcal{E}_b^{\mathrm{valid}}
    \quad \Longleftrightarrow \quad
    m_{b,i}=1 \ \text{and}\ m_{b,j}=1.
\end{equation}
Thus, the base topology is identical for all participants, but the effective topology is participant-specific. If a participant \(b\) has \(d_b\) valid days, the effective graph contains \(d_b^2\) directed edges, including self-loops. For a fully observed participant, this recovers the full 441-edge graph; for partially observed participants, the graph remains calendar-aligned but excludes invalid-day communication. This design prevents missing days from sending or receiving messages while preserving the full calendar structure.

Node features are first projected to a 64-dimensional latent representation using a linear projection, layer normalization, GELU activation, and one-dimensional spatial dropout with probability \(p=0.2\). Two residual GATv2 \cite{brody2022gatv2} layers are then applied, each with four attention heads and 16 hidden dimensions per head. The GATv2 layers receive the temporal edge attributes directly. Additional self-loops are not inserted because self-connections are already included in the base graph and explicitly encoded as an edge attribute.

A learnable distance-decay channel is appended to the edge attributes to discourage unsupported long-range temporal routing while allowing the strength of this temporal bias to be learned from data:
\begin{equation}
    b_{ij} = -\operatorname{softplus}(\alpha)\frac{|i-j|}{20},
\end{equation}
where \(\alpha\) is a trainable scalar. The effective edge feature vector is therefore six-dimensional. Residual propagation for layer \(\ell\) is implemented as
\begin{equation}
\begin{aligned}
\widetilde{\mathbf{H}}^{(\ell+1)}
&= \operatorname{Drop}\!\left(
\operatorname{GELU}\!\left(
\operatorname{LN}\!\left(
\operatorname{GATv2}_{\ell}
(\mathbf{H}^{(\ell)}, \mathcal{E}^{\mathrm{valid}}, \mathbf{E})
\right)\right)\right), \\
\mathbf{H}^{(\ell+1)}
&= \mathbf{H}^{(\ell)} + \widetilde{\mathbf{H}}^{(\ell+1)} .
\end{aligned}
\label{eq:residual_gat}
\end{equation}
Here, \(\mathcal{E}^{\mathrm{valid}}\) denotes the dynamically pruned edge set, \(\mathbf{E}\) denotes the corresponding edge attributes after appending the distance-decay channel, and \(\widetilde{\mathbf{H}}^{(\ell+1)}\) denotes the residual update produced by the \(\ell\)-th GATv2 layer.

The graph-level wearable embedding is obtained by masked attention pooling. A trainable query vector \(\mathbf{q}\in\mathbb{R}^{64}\) computes an attention score for each day. Invalid nodes are assigned \(-\infty\) before softmax normalization:
\begin{equation}
s_t =
\begin{cases}
\mathbf{h}_t \mathbf{q}^{\top}, & m_t = 1,\\
-\infty, & m_t = 0,
\end{cases}
\label{eq:pool_score}
\end{equation}
\begin{equation}
w_t = \operatorname{softmax}(s_t), \qquad
\mathbf{z}_{\mathrm{wear}} = \sum_{t=1}^{21} w_t \mathbf{h}_t .
\label{eq:masked_pool}
\end{equation}

The continuous baseline covariates (age and body mass index) were standardized within each training fold. The categorical covariates (biological sex, study site, highest parental education, and combined household income) were one-hot encoded, with unseen validation or test categories ignored during transformation. The encoded covariates were passed through a 32-dimensional multilayer perceptron branch and concatenated with the graph-level wearable embedding. The fused representation was then passed through a 64-dimensional regression head to predict $G_{c}$. Importantly, this architecture natively addresses the multicollinearity inherent in 24-hour time allocation variables. While traditional linear models are highly sensitive to these correlated temporal predictors, SATURN's dense projection and graph attention layers map them into a shared latent space prior to fusion, mitigating collinearity while preserving their joint behavioral signal.

\subsection{Training and Evaluation Protocol}

All data-dependent preprocessing transformations were fit on the training split within each fold, then applied to the corresponding validation and test splits. The target score was transformed to approximately normality using a quantile transformation fitted only to the training targets. Model predictions were inverse-transformed to the original score scale before computing mean squared error, mean absolute error, Pearson correlation, and coefficient of determination on validation and test subjects.

The graph model was trained with AdamW using a learning rate of \(10^{-3}\) and weight decay of \(10^{-4}\). Dropout was used as regularization within the graph and regression components. The mini-batch size was 128, the maximum number of epochs was 300, and gradients were clipped to a maximum norm of 1.0. The training objective was mean squared error in the transformed target space. A reduce-on-plateau learning-rate scheduler monitored validation \(R^2\), reduced the learning rate by a factor of 0.5 after 10 stagnant epochs, and used a minimum learning rate of \(10^{-6}\). Early stopping was based on validation \(R^2\) with a patience of 30 epochs, and the best-validation checkpoint was evaluated once on the held-out test fold. 

\subsection{Explainable AI and Uncertainty Quantification}

Model interpretation was conducted on the held-out test subjects of each fold using the saved best checkpoints. Permutation feature importance measured the decrease in test \(R^2\) after shuffling a feature group across subjects \cite{fisher2019allmodels}. For temporal wearable features, the full 21-day trajectory of each feature group was permuted across subjects, preserving within-feature temporal shape while breaking subject-feature correspondence. Cyclic sine-cosine pairs and one-hot week indicators were treated as grouped features. For categorical covariates, all one-hot columns belonging to the same covariate were permuted together.

Integrated Gradients was computed with 50 interpolation steps \cite{sundararajan2017ig}. The baseline set standardized biological temporal features and covariates to zero, while preserving the graph-validity mask throughout the integration path. This prevented the baseline from changing graph topology and isolated attribution to biological and sociodemographic inputs. Absolute Integrated Gradients magnitudes were averaged across subjects and folds, with grouped attributions summed for cyclic and one-hot feature groups. For explainability convergence analysis, structural mask variables, temporal index variables, and device-compliance duration variables were excluded, allowing rank correlation to focus on interpretable behavioral and covariate predictors.

To visualize learned temporal routing, final-layer GATv2 attention weights were extracted from each trained model. Attention values were averaged over the four attention heads, projected from the sparse participant-specific edge lists back to dense \(21\times21\) matrices, and averaged across all test subjects and folds. These attention maps are interpreted as diagnostics of learned message routing rather than causal explanations. 

Epistemic uncertainty was estimated using Monte Carlo dropout \cite{gal2016dropout}: dropout and spatial-dropout modules were reactivated at inference, and 100 stochastic forward passes were run per test subject. This number of passes was selected to provide a stable empirical estimate of epistemic uncertainty while remaining computationally efficient during inference. The mean and standard deviation of inverse-transformed predictions were used as the point prediction and uncertainty proxy, respectively.

\section{Results and Discussion}

\subsection{Covariate Baseline and Demographic Signal}
The covariate-only experiments quantify the predictive information available from age, sex, BMI, parental education, household income, and study site before adding Fitbit telemetry. Hyperparameter tuning for traditional machine learning models used the predefined validation split used in the deep learning experiments, keeping the train-validation-test structure aligned across model families.

As shown in Table \ref{tab:covariates_baseline}, the covariate-only baselines were strong. Lasso regression achieved the highest covariate-only performance ($R^2=0.2346\pm0.0205$, $r=0.4868\pm0.0228$), with Ridge and ElasticNet performing similarly. This result is expected for $G_c$, which is strongly related to developmental exposure, education-linked household context, and age. Consequently, the appropriate question for the wearable models is not whether Fitbit data alone can replace these covariates but whether activity--sleep sequences add reproducible predictive information beyond them.

\begin{table*}[t]
\caption{Baseline Predictive Performance for Crystallized Intelligence (\bm{$G_c$}) Using Covariates Only.}
\label{tab:covariates_baseline}
\centering
\renewcommand{\arraystretch}{1.2}
\begin{tabular*}{\textwidth}{@{\extracolsep{\fill}} l c c c c}
\toprule
\textbf{Method} & \textbf{\bm{$R^2 \uparrow$}} & \textbf{r $\bm{\uparrow}$} & \textbf{MAE $\bm{\downarrow}$} & \textbf{MSE $\bm{\downarrow}$} \\
\midrule
Ridge Regression & 0.2345 $\pm$ 0.0217 & 0.4866 $\pm$ 0.0240 & 4.7687 $\pm$ 0.1114 & 36.8723 $\pm$ 1.3702 \\
Lasso Regression & \textbf{0.2346 $\pm$ 0.0205} & \textbf{0.4868 $\pm$ 0.0228} & \textbf{4.7625 $\pm$ 0.1131} & \textbf{36.8663 $\pm$ 1.3438} \\
ElasticNet & 0.2340 $\pm$ 0.0230 & 0.4858 $\pm$ 0.0253 & 4.7676 $\pm$ 0.1174 & 36.8966 $\pm$ 1.4225 \\
Random Forest & 0.1964 $\pm$ 0.0226 & 0.4462 $\pm$ 0.0239 & 4.8769 $\pm$ 0.1326 & 38.7122 $\pm$ 1.6239 \\
Gradient Boosting & 0.2207 $\pm$ 0.0128 & 0.4727 $\pm$ 0.0146 & 4.7996 $\pm$ 0.1106 & 37.5413 $\pm$ 1.3297 \\
Support Vector Regression & 0.2071 $\pm$ 0.0196 & 0.4633 $\pm$ 0.0263 & 4.8371 $\pm$ 0.1140 & 38.1910 $\pm$ 1.3762 \\
\bottomrule
\end{tabular*}

\vspace{2mm}
\parbox{1\textwidth}{\small \textit{Note:} Bold indicates the best performing model. \(\uparrow/\downarrow\): higher/lower is better. Lasso regression provided the strongest covariate-only baseline in this experiment; wearable models should be interpreted relative to this demographic and site-adjusted benchmark.}
\end{table*}


\subsection{Multimodal Model Comparison and Sequence Topologies}
Table~\ref{tab:method_comparison} compares three modeling regimes: flattened machine learning, sequential recurrent learning, and the proposed temporal graph model. All models were evaluated on the same family-aware cross-validation folds using identical covariate sets unless otherwise noted.

Flattened models treat the 21-day sequence as independent columns in a tabular dataset. Despite improving upon simple linear baselines, the best flattened model (Gradient Boosting, $R^2 = 0.2372 \pm 0.0136$) only marginally exceeded the strongest covariate-only result ($R^2 = 0.2346$). This suggests that simple concatenation fails to preserve adjacent-day continuity or to handle intermittent missingness. Conversely, the bidirectional LSTM (BiLSTM) with masked mean pooling processed sequences chronologically, successfully exceeding demographic baselines ($R^2 = 0.2688 \pm 0.0154$). However, substituting the masking mechanism with conventional mean or forward-fill imputation degraded BiLSTM performance. This indicates that forcing sequential models to process synthetically filled gaps propagates stale states or population-average noise.

The proposed SATURN architecture achieved the highest overall performance ($R^2 = 0.2783 \pm 0.0127$, $r = 0.5304 \pm 0.0104$). By treating missingness as a structural property rather than a sequential gap, SATURN dynamically prunes invalid nodes and employs a learnable distance-decay channel. Ablating this dynamic topology by feeding imputed data into the GATv2 layers reduced performance to $R^2 = 0.2735$ (mean imputation), confirming that topological pruning is highly effective for the irregular adherence patterns inherent to free-living wearable data. 

Finally, evaluating \modelname{} without sociodemographic covariates yielded $R^2 = 0.1221 \pm 0.0088$, verifying the presence of an independent behavioral signal. Crucially, the performance of the fully fused network ($R^2 = 0.2783$) is substantially less than the simple arithmetic sum of the independent covariate baseline ($R^2 = 0.2346$) and the wearable-only model. This sub-additive variance overlap provides compelling empirical validation of a core tenet in computational social systems: an adolescent's free-living movement behavior, sleep hygiene, and device compliance patterns are fundamentally bounded and scaffolded by their sociodemographic reality. Rather than rendering the streams redundant, \modelname{} successfully models how this socioeconomic framework manifests physically in daily life, extracting a modest but highly reproducible incremental behavioral signal layered on top of strong social structure.


\begin{table*}[t]
\caption{Comparison of Predictive Performance for Crystallized Intelligence (\bm{$G_c$}) Across Methodological Families.}
\label{tab:method_comparison}
\centering
\renewcommand{\arraystretch}{1.2}
\begin{tabular*}{\textwidth}{@{\extracolsep{\fill}} l c c c c}
\toprule
\textbf{Method Configuration} & \textbf{\bm{$R^2 \uparrow$}} & \textbf{r $\bm{\uparrow}$} & \textbf{MAE $\bm{\downarrow}$} & \textbf{MSE $\bm{\downarrow}$} \\
\midrule
\multicolumn{5}{l}{\textbf{Flattened Machine Learning (Zero-Padding)}} \\
\quad \; -- Ridge Regression & 0.2171 $\pm$ 0.0135 & 0.4696 $\pm$ 0.0132 & 4.8077 $\pm$ 0.1270 & 37.7280 $\pm$ 1.6871 \\
\quad \; -- Lasso Regression & 0.2287 $\pm$ 0.0121 & 0.4825 $\pm$ 0.0142 & 4.7788 $\pm$ 0.1177 & 37.1644 $\pm$ 1.4363 \\
\quad \; -- ElasticNet & 0.2368 $\pm$ 0.0095 & 0.4897 $\pm$ 0.0111 & \textbf{4.7434 $\pm$ 0.1094} & 36.7727 $\pm$ 1.3978 \\
\quad \; -- Random Forest & 0.2071 $\pm$ 0.0104 & 0.4622 $\pm$ 0.0140 & 4.8461 $\pm$ 0.1279 & 38.2130 $\pm$ 1.6671 \\
\quad \; -- Gradient Boosting & \textbf{0.2372 $\pm$ 0.0136} & \textbf{0.4902 $\pm$ 0.0150} & 4.7523 $\pm$ 0.1001 & \textbf{36.7395 $\pm$ 1.1104} \\
\quad \; -- Support Vector Regression & 0.1713 $\pm$ 0.0139 & 0.4157 $\pm$ 0.0156 & 4.9420 $\pm$ 0.1310 & 39.9211 $\pm$ 1.4254 \\
\midrule
\multicolumn{5}{l}{\textbf{Sequential Deep Learning (BiLSTM)}} \\
\quad \; -- Masked Mean Pooling & \textbf{0.2688 $\pm$ 0.0154} & \textbf{0.5215 $\pm$ 0.0135} & \textbf{4.6532 $\pm$ 0.1223} & \textbf{35.2200 $\pm$ 1.1794} \\
\quad \; -- Mean Imputation & 0.2655 $\pm$ 0.0153 & 0.5187 $\pm$ 0.0141 & 4.6695 $\pm$ 0.1183 & 35.3828 $\pm$ 1.2682 \\
\quad \; -- Forward-Fill (LOCF) & 0.2647 $\pm$ 0.0155 & 0.5171 $\pm$ 0.0131 & 4.6612 $\pm$ 0.1008 & 35.4129 $\pm$ 0.9672 \\
\midrule
\multicolumn{5}{l}{\textbf{Graph Deep Learning (SATURN)}} \\
\quad \; -- Dynamic Edge Pruning (Proposed) & \textbf{0.2783 $\pm$ 0.0127} & \textbf{0.5304 $\pm$ 0.0104} & \textbf{4.6180 $\pm$ 0.0916} & \textbf{34.7568 $\pm$ 0.8795} \\
\quad \; -- Mean Imputation & 0.2735 $\pm$ 0.0125 & 0.5245 $\pm$ 0.0117 & 4.6367 $\pm$ 0.1061 & 34.9921 $\pm$ 1.0888 \\
\quad \; -- Forward-Fill (LOCF) & 0.2727 $\pm$ 0.0143 & 0.5241 $\pm$ 0.0147 & 4.6406 $\pm$ 0.1118 & 35.0290 $\pm$ 1.1029 \\
\quad \; -- without Covariates (Fitbit Only) & 0.1221 $\pm$ 0.0088 & 0.3548 $\pm$ 0.0094 & 5.1126 $\pm$ 0.1205 & 42.2957 $\pm$ 1.5398 \\
\bottomrule
\end{tabular*}

\vspace{2mm}
\parbox{1\textwidth}{\small \textit{Note:} Bold indicates the best performing model within its methodological family. \(\uparrow/\downarrow\): higher/lower is better. All models utilize the combined Fitbit and sociodemographic covariate feature set unless explicitly noted otherwise. Flattened ML models treat the 21-day sequence as independent tabular features, whereas sequential and graph models process the temporal structure natively.}
\end{table*}

\begin{figure*}[!t]
    \centering
    \includegraphics[width=\textwidth]{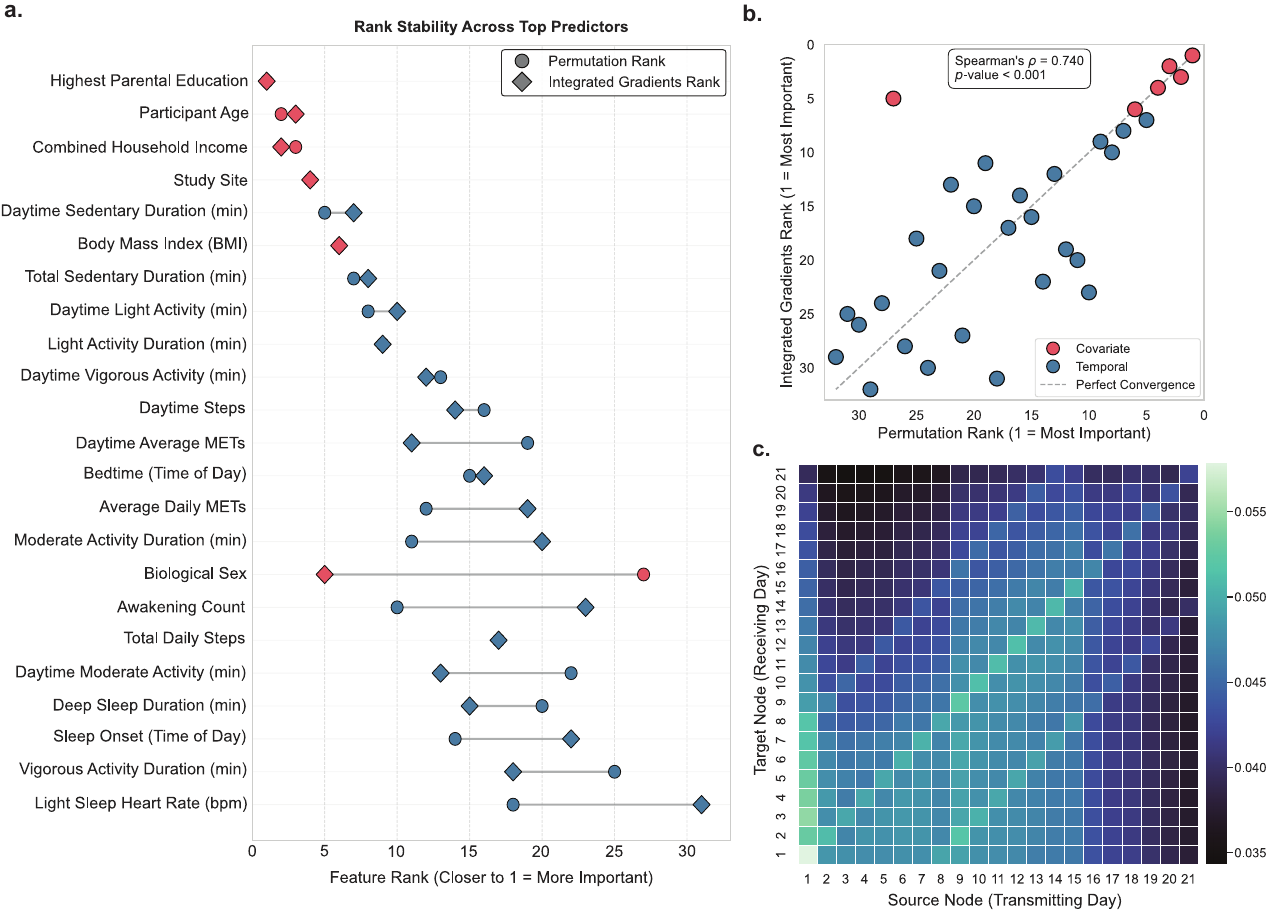}
    \caption{Explainable AI (XAI) and temporal attention diagnostics for the \modelname{} architecture. Panel (a) shows top-predictor rank stability between Permutation Feature Importance (circle) and Integrated Gradients (diamond), with connecting segments showing method disagreement. Panel (b) shows global rank convergence between the two XAI methods (Spearman's $\rho=0.740$, $p<0.001$), separating covariates from temporal wearable features. Panel (c) visualizes the mean final-layer $21\times21$ edge-attention matrix, where rows denote receiving days and columns denote transmitting days.}
    \label{fig:xai}
\end{figure*}

\subsection{Explainable AI Convergence and Behavioral Interpretation}
PFI and IG provided complementary views of model reliance. PFI estimates the functional performance drop caused by disrupting a feature group, whereas IG measures local gradient-based attribution along a baseline-to-input path. After excluding structural masks, temporal index variables, and device-compliance variables from the convergence plot, the two attribution methods showed substantial rank agreement (Spearman's $\rho=0.740$, $p<0.001$). This agreement supports the stability of the main behavioral interpretation, although PFI and IG should not be expected to produce identical scales because they answer different questions.

The most consistent temporal predictors included light activity duration, average daily METs, total sleep-period measures, and sedentary duration. Panel (a) in Fig. \ref{fig:xai} further shows that the strongest covariate signals were parental education, age, household income, and study site, which is consistent with the high covariate-only baseline. This result is expected for $G_c$, which is heavily shaped by cumulative environmental exposure, formal education, and age \cite{luciana2018abcd, loughnan2023intelligence}. Notably, the XAI analysis identified the highest parental education as the dominant demographic predictor, aligning with recent ABCD cohort studies that highlight the outsized role of caregiver education and household socioeconomic status in scaffolding crystallized cognitive development \cite{marzoratti2026contextualizing}. Therefore, this should be interpreted as an important contextual finding rather than a nuisance signal: wearable features should be interpreted as an incremental behavioral signal layered on top of strong social and demographic structure.

The rank scatter in Fig. \ref{fig:xai}(b) is also informative. Covariates concentrated near the highest ranks for both methods, whereas temporal features showed broader dispersion, indicating that the two XAI methods agree on the broad signal hierarchy but differ on some feature-specific ordering. Such disagreement is expected because PFI asks how much test performance drops when a feature group is disrupted, whereas IG attributes local prediction sensitivity along a baseline-to-input path. Therefore, features that are redundant with correlated variables may receive lower PFI scores than their gradient sensitivity would suggest, and vice versa.

The attention heatmap in Fig. \ref{fig:xai}(c) provides an additional diagnostic of temporal routing. The final GATv2 layer placed stronger mass within local chronological neighborhoods and showed broader off-diagonal structure across the 21-day window, consistent with the model's use of both short-range continuity and repeated weekly context. Because attention weights are normalized within the graph attention operation and are affected by feature values, edge attributes, and valid-day masks, they should be read as learned routing patterns rather than standalone feature importance scores.

\begin{figure*}[!t]
    \centering
    \includegraphics[width=\textwidth]{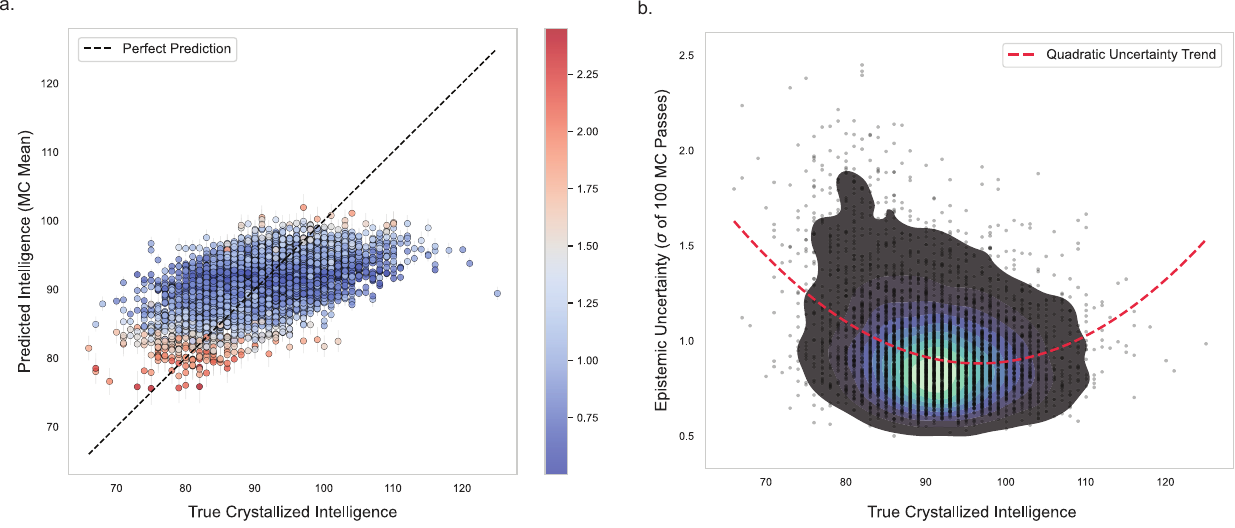}
    \caption{Epistemic uncertainty estimation via 100 Monte Carlo dropout passes. Panel (a) compares true crystallized intelligence scores with the Monte Carlo predictive mean; vertical error bars and point color indicate the predictive standard deviation. Panel (b) plots true score against epistemic uncertainty using a bivariate density overlay and a quadratic trend line, showing lower uncertainty near the densely sampled population center and higher uncertainty toward the score tails.}
    \label{fig:uncertainty}
\end{figure*}

\subsection{MC Dropout Epistemic Uncertainty Mapping}
Figure \ref{fig:uncertainty} summarizes epistemic uncertainty estimated from 100 Monte Carlo dropout passes. Panel (a) shows the expected regression-to-the-mean pattern: lower true scores tend to be overestimated and higher true scores tend to be underestimated. This does not invalidate the predictive result, but it is important for interpretation because the model is more reliable for ranking and central-range estimation than for precise prediction of extreme $G_c$ values. Panel (b) shows that predictive standard deviation increased toward the lower and upper tails of $G_c$, consistent with the lower density of extreme target values and the tendency of regression models to be most stable near the distribution center.

For digital phenotyping, this uncertainty profile is useful because it discourages overconfident interpretation of atypical cases. If wearable-based models are considered for screening or monitoring, high-uncertainty predictions should be flagged for additional assessment rather than treating all predicted scores as equally reliable. The present uncertainty estimates are nevertheless approximate: MC dropout captures one source of epistemic uncertainty, but it does not fully calibrate predictive intervals or account for unmeasured confounding. Future research should incorporate calibration diagnostics, prediction-interval coverage, and subgroup-specific uncertainty to establish whether epistemic confidence remains equally reliable across diverse demographic strata.

\subsection{Subgroup Robustness and Calibration}

Table~\ref{tab:subgroup_fairness} summarizes SATURN's pooled held-out subgroup performance across the five family-aware folds. Overall pooled performance was consistent with the fold-averaged model comparison result, with \(R^2=0.279\), \(r=0.529\), MAE \(=4.62\), and RMSE \(=5.90\). Notably, sex-based differences were minimal: females and males showed similar absolute error, with MAE values of 4.60 and 4.64, respectively, and small calibration biases, indicating that the model performs equitably across biological sex.

Socioeconomic subgroups showed clearer differences in the underlying \(G_c\) distributions. Mean \(G_c\) increased from 83.66 in the low-parental-education group to 93.04 in the high parental-education group, and from 85.78 in the low-income group to 93.16 in the high-income group. Despite these distributional differences, the absolute error remained broadly comparable in interpretable education and income strata. MAE ranged from 4.56 to 4.76 in parental-education groups and from 4.52 to 4.83 in household-income groups. The largest calibration difference was observed for the low parental-education group, where \modelname{} over-predicted by 0.95 points on average, accompanied by a pronounced reduction in within-group variance explained \(R^2=0.029\). Although this drop in \Rtwo {} is partially a function of the restricted variance of the target within the group and a smaller sample size (N=280), it fundamentally highlights a critical boundary condition in digital phenotyping. Deep learning architectures inherently optimize for global data density, leaving minoritized distributions vulnerable to higher relative epistemic uncertainty. From a sociotechnical perspective, this decoupling suggests that in lower-resource environments, the systemic day-to-day physical behaviors captured by commercial wearables may share less direct mutual information with standardized, formal academic testing metrics due to external environmental constraints. Consequently, absolute error and calibration bias must remain the primary benchmarks for assessing equitable cross-strata utility, rather than variance-dependent metrics alone. Participants with fewer than 21 valid wearable days did not show degraded performance relative to the high-adherence group, supporting the robustness of the missingness-aware graph design.


\begin{table*}[t]
\caption{Subgroup Robustness and Calibration Analysis for SATURN (Pooled Held-Out Predictions)}
\label{tab:subgroup_fairness}
\centering
\scriptsize
\renewcommand{\arraystretch}{1.1}
\begin{tabular*}{\textwidth}{@{\extracolsep{\fill}} l r c c c c c c}
\toprule
\textbf{Subgroup} & \textbf{N} & \textbf{\(\bm{G_c}\) Mean \(\pm\) SD [Range]} & \textbf{Bias} & \textbf{MAE \(\bm{\downarrow}\)} & \textbf{RMSE \(\bm{\downarrow}\)} & \textbf{r \(\bm{\uparrow}\)} & \textbf{\(\bm{R^2 \uparrow}\)} \\
\midrule
\textbf{Overall Cohort} & 6091 & 91.26 $\pm$ 6.94 [66--125] & -0.19 & 4.62 & 5.90 & 0.529 & 0.279 \\
\midrule
\textbf{Biological Sex} \\
\; -- Female & 2957 & 91.10 $\pm$ 7.00 [66--121] & -0.21 & 4.60 & 5.87 & 0.546 & 0.297 \\
\; -- Male & 3134 & 91.41 $\pm$ 6.89 [67--125] & -0.18 & 4.64 & 5.92 & 0.512 & 0.261 \\
\midrule
\textbf{Parental Education} \\
\; -- Low ($\leq$12 yrs) & 280 & 83.66 $\pm$ 6.15 [66--104] & 0.95 & 4.76 & 6.05 & 0.324 & 0.029 \\
\; -- Medium (13--16 yrs) & 1896 & 88.73 $\pm$ 6.54 [67--125] & -0.13 & 4.56 & 5.79 & 0.463 & 0.214 \\
\; -- High ($\geq$17 yrs) & 3905 & 93.04 $\pm$ 6.42 [70--121] & -0.31 & 4.63 & 5.93 & 0.386 & 0.146 \\
\midrule
\textbf{Household Income} \\
\; -- Low ($<$\$25k) & 585 & 85.78 $\pm$ 6.66 [66--111] & 0.02 & 4.55 & 5.84 & 0.482 & 0.230 \\
\; -- Medium (\$25k--\$75k) & 1612 & 89.76 $\pm$ 6.71 [68--116] & -0.01 & 4.83 & 6.10 & 0.419 & 0.174 \\
\; -- High ($\geq$\$75k) & 3488 & 93.16 $\pm$ 6.28 [70--121] & -0.31 & 4.52 & 5.79 & 0.393 & 0.152 \\
\; -- Unknown & 406 & 88.84 $\pm$ 7.27 [67--125] & -0.26 & 4.76 & 6.09 & 0.549 & 0.298 \\
\midrule
\textbf{Wearable Adherence} \\
\; -- Low ($<$21 valid days) & 2345 & 90.09 $\pm$ 7.08 [66--118] & -0.23 & 4.58 & 5.85 & 0.564 & 0.317 \\
\; -- High ($\geq$21 valid days) & 3746 & 91.99 $\pm$ 6.76 [68--125] & -0.17 & 4.64 & 5.93 & 0.482 & 0.232 \\
\bottomrule
\end{tabular*}

\vspace{1mm}
\parbox{1\textwidth}{\small \textit{Note:} Bias is predicted minus observed \(G_c\) (closer to zero indicates better calibration). Subgroup \(R^2\) depends heavily on within-group target variance. The unknown parental-education group ($N=10$) is omitted.}
\end{table*}

\subsection{Limitations and Implications for Computational Social Systems}
Several limitations should guide interpretation. First, the analysis was cross-sectional, intentionally conducted at the two-year follow-up to maximize the density of concurrent, high-quality wearable compliance and $G_c$ assessments. This serves as a necessary foundational validation of the temporal graph approach, though it supports predictive association rather than causal inference. Future longitudinal trajectory modeling is required. Second, the cohort was drawn from United States adolescents in the ABCD Study, and external validation is required before generalizing the model to other countries, age ranges, wearable devices, or data-collection protocols. Third, the analyses used tabulated daily summaries rather than raw minute-level sensor streams, so the model could not capture within-day activity timing, bout structure, or fine-grained sleep-stage dynamics beyond the available daily aggregates. Fourth, the target was limited to uncorrected $G_c$ at the two-year follow-up; future work should evaluate fluid and total cognition, longitudinal change, and multimodal fusion with neuroimaging or questionnaire-derived behavioral measures.

A further limitation concerns missingness in wearable data and subgroup robustness. Consumer wearable data are shaped not only by behavior, but also by device adherence, family context, socioeconomic opportunity, and participation patterns. Prior ABCD wearable analyses have shown that wearable-device participation and wear time vary across demographic and socioeconomic groups \cite{kim2023jama}, and fairness concerns remain under-addressed in mobile and wearable machine-learning research \cite{yfantidou2026unfair}. SATURN explicitly addresses this missingness through dynamic graph masking (as detailed in Section II.E). While unobserved days are zero-padded to maintain consistent sequence lengths, the validity mask topologically isolates these invalid nodes by severing their message-passing edges and forcing their global attention weights to zero. This ensures the model learns strictly from observed behavior rather than imputation artifacts. However, missingness itself may still encode social structure \cite{kiang2021sociodemographic, currey2023increasing}, and extensive edge pruning for highly non-compliant participants reduces the graph to a sparse state, potentially resulting in the loss of critical temporal resolution. Future work should therefore report performance, calibration, and uncertainty across sex, parental education, household income, study site, and wearable-adherence strata to determine whether the incremental wearable signal is equally reliable across subgroups.

Despite these constraints, the study is relevant to computational social systems because it models passive human--device traces without collapsing temporal structure or treating missingness as simple noise. The proposed framework links behavioral regularity, sociodemographic context, and machine learning methodology: covariates define a strong demographic benchmark, while the temporal graph captures incremental activity--sleep signal under explicit validity constraints. This provides a reproducible framework for studying wearable-derived behavioral phenotypes in adolescent populations, where observed data are jointly shaped by individual behavior, social context, and technology-mediated measurement.

\section{Conclusion}
Free-living wearable data is inherently incomplete, and treating that incompleteness as a modeling nuisance rather than a structural property of the signal is the core failure of prior approaches. SATURN resolves this by pruning invalid days within the temporal graph rather than imputing them, recovering a behavioral hierarchy (\textit{e.g.}, light physical activity, metabolic equivalents, and sleep duration) that predicts adolescent $G_c$ above sociodemographic baselines. Subgroup and fairness analyses confirm that this advantage is not concentrated in demographically convenient subgroups, as absolute error remains consistent across household income, parental education, and adherence strata. Monte Carlo uncertainty estimates further reveal that epistemic confidence tracks data density, providing a principled deployment signal rather than uniform overconfidence.
Taken together, the results establish missingness-aware temporal graphs as a rigorous substrate for continuous digital cognitive phenotyping, one whose validity derives not from architectural complexity but from respect for the sociotechnical structure of the data it models.

Practically, this framework provides a scalable, ecologically valid computational substrate to identify macro-level deviations in expected cognitive-behavioral trajectories. By highlighting these behavioral anomalies passively, it offers a proactive stratification mechanism to help researchers and clinicians prioritize resource allocation for traditional, in-depth clinical neurocognitive assessments.

\section*{Acknowledgment}
This research was supported by the Science and Technology Fellowship Trust, Government of the People's Republic of Bangladesh, and the Commonwealth through an Australian Government Research Training Program Scholarship. Data used in the preparation of this article were obtained from the Adolescent Brain Cognitive Development (ABCD) Study, held in the NIH Brain Development Cohorts Data Sharing Platform. This work was also supported by resources provided by the University of Queensland Research Computing Center's Bunya supercomputer.


\printbibliography
\end{document}